%% file: main.tex
\documentclass[letterpaper]{article} 
\usepackage{aaai2026}
\usepackage{times}  
\usepackage{helvet}  
\usepackage{courier}  
\usepackage[hyphens]{url}  
\usepackage{graphicx} 
\usepackage{multirow}
\usepackage{booktabs}
\usepackage{amssymb}
\usepackage{makecell}
\usepackage{xcolor}
\usepackage{cuted}
\usepackage{amsmath}
\nocopyright

\usepackage{natbib}  
\usepackage{caption} 
\usepackage{algorithm}
\usepackage{algorithmic}

\usepackage{newfloat}
\usepackage{listings}
\DeclareCaptionStyle{ruled}{labelfont=normalfont,labelsep=colon,strut=off} 
\floatstyle{ruled}
\newfloat{listing}{tb}{lst}{}
\floatname{listing}{Listing}
\title{SeqAffordSplat: Scene-level Sequential Affordance Reasoning on 3D Gaussian Splatting}

\author{
    Di Li{\rm 1}, Jie Feng{\rm 1}\thanks{Corresponding author.}, Jiahao Chen{\rm 2}, 
    Weisheng Dong{\rm 1}, \\Guanbin Li{\rm 2}, Yuhui Zheng{\rm 3},Mingtao Feng{\rm 1}, Guangming Shi{\rm 1}}
\affiliations{
    \textsuperscript{\rm 1}Xidian University\\
    \textsuperscript{\rm 2}Sun Yat-sen University\\
    \textsuperscript{\rm 3}Qinghai Normal University\\
    dili@stu.xidian.edu.cn, jiefeng0109@163.com,chenjh328@mail2.sysu.edu.cn, wsdong@mail.xidian.edu.cn, \\
    liguanbin@mail.sysu.edu.cn, zhengyh@vip.126.com, mintfeng@hnu.edu.cn, gmshi@xidian.edu.cn
    
}

\begin{document}

\maketitle

\begin{strip}
  \centering
  \includegraphics[width=\linewidth]{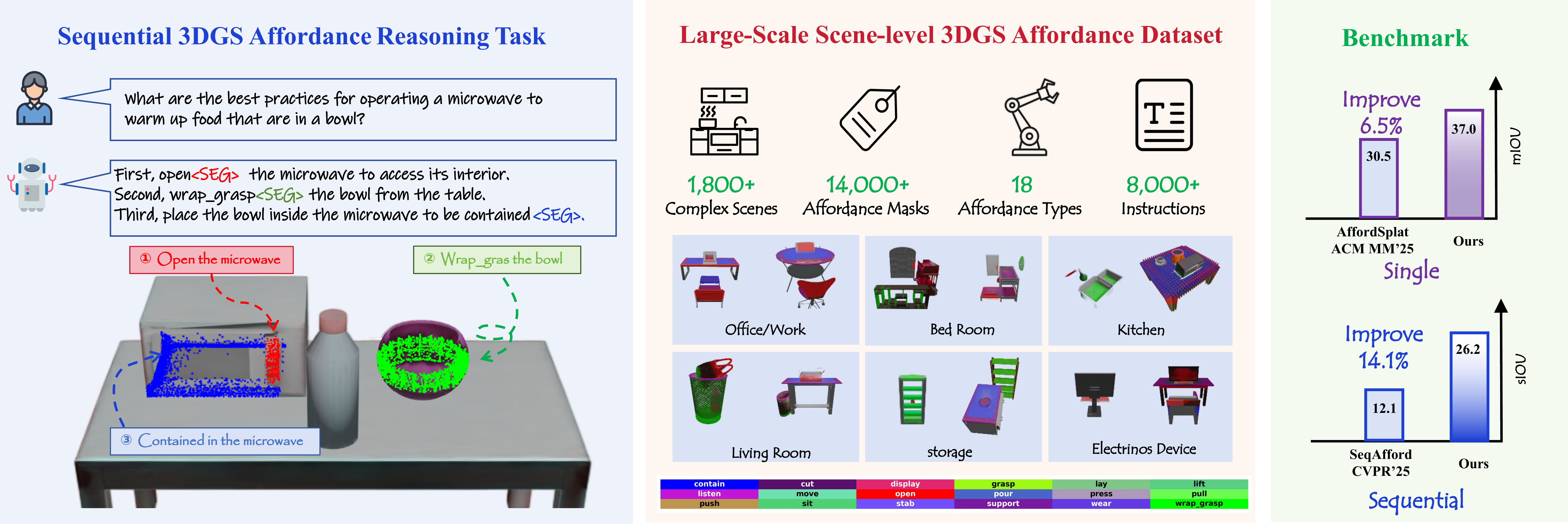}
  \captionof{figure}{(\textbf{Left})We introduce \textbf{Sequential 3DGS Affordance Reasoning Task} for complex, multi-step agent interactions. (\textbf{Center})To support this, we present \textbf{SeqAffordSplat}, a large-scale dataset with over 1,700 3DGS scenes and 12,000 instruction pairs.(\textbf{Right}) Our model, \textbf{SeqSplatNet}, sets a new state-of-the-art, improving performance by 6.5\% on single-step tasks and 14.1\% on our sequential benchmark. Please zoom in for better visual effects.}
  \label{fig:teaser}
\end{strip}

\begin{abstract}
3D affordance reasoning, the task of associating human instructions with the functional regions of 3D objects, is a critical capability for embodied agents. Current methods based on 3D Gaussian Splatting (3DGS) are fundamentally limited to single-object, single-step interactions, a paradigm that falls short of addressing the long-horizon, multi-object tasks required for complex real-world applications. To bridge this gap, we introduce the novel task of Sequential 3D Gaussian Affordance Reasoning and establish SeqAffordSplat, a large-scale benchmark featuring 1800+ scenes to support research on long-horizon affordance understanding in complex 3DGS environments. We then propose SeqSplatNet, an end-to-end framework that directly maps an instruction to a sequence of 3D affordance masks. SeqSplatNet employs a large language model that autoregressively generates text interleaved with special segmentation tokens, guiding a conditional decoder to produce the corresponding 3D mask. To handle complex scene geometry, we introduce a pre-training strategy, Conditional Geometric Reconstruction, where the model learns to reconstruct complete affordance region masks from known geometric observations, thereby building a robust geometric prior. Furthermore, to resolve semantic ambiguities, we design a feature injection mechanism that lifts rich semantic features from 2D Vision Foundation Models (VFM) and fuses them into the 3D decoder at multiple scales. Extensive experiments demonstrate that our method sets a new state-of-the-art on our challenging benchmark, effectively advancing affordance reasoning from single-step interactions to complex, sequential tasks at the scene level.
\end{abstract}

\begin{table*}[htbp]
  \centering
  \small
  \setlength{\tabcolsep}{4pt}
  \caption{Comparison of Existing 3D Affordance Datasets with Ours.}
  \label{tab:dataset}
  \begin{tabular}{lccccc}
    \toprule
    \textbf{Benchmark} & \makecell{\textbf{Vision} \\ \textbf{Type}} & \makecell{\textbf{Scene-level} \\ \textbf{Support}} & \makecell{\textbf{Sequence} \\ \textbf{Support}} & \makecell{\textbf{\#Object} \\ \textbf{Cat.}} & \makecell{\textbf{\#Afford.} \\ \textbf{Type}} \\
    \midrule
    3D AffordanceNet (CVPR'21) & PointCloud & $\times$ & $\times$ & 23 & 17 \\
    LASO (CVPR'24) & PointCloud & $\times$ & $\times$ & 23 & 17 \\
    SeqAfford (CVPR'25) & PointCloud & $\times$ & \checkmark & 23 & 18 \\
    \midrule
    3DAffordSplat (ACM MM'25) & Gaussians & $\times$ & $\times$ & 21 & 18 \\
    \textbf{Ours} & \textbf{Gaussians} & \textbf{\checkmark} & \textbf{\checkmark} & \textbf{21} & \textbf{18} \\
    \bottomrule
  \end{tabular}
\end{table*}

\section{Introduction}

3D affordance reasoning, which identifies interactive regions on objects to enable specific actions in 3D space, is a fundamental perceptual capability for embodied agents~\cite{deng20213d,yang2023grounding}.
By linking perception and action, it underpins essential functionalities in a spectrum of applications, including robotic manipulation~\cite{yamanobe2017brief}, augmented reality~\cite{steffen2019framework,nagarajan2020learning}, and virtual reality~\cite{dalgarno2010learning,venkatakrishnan2023virtual}. 
This has motivated early explorations into affordance reasoning using point cloud representations~\cite{yang2023grounding,li2024laso,deng20213d}.
While these approaches demonstrate potential in predicting affordances from 3D geometry, they are often constrained by the inherent sparsity and discrete nature of point clouds, which impedes their ability to capture the fine-grained, continuous structures essential for precise interaction.

Inspired by the high-fidelity representations of 3D Gaussian Splatting (3DGS)~\cite{kerbl20233d}, the transition from sparse point clouds to 3DGS has drawn increasing attention for 3D scene understanding~\cite{mohammadi20233dsgrasp,zhu20243d}.  
The pioneering work~\cite{wei20253daffordsplat} demonstrates promising improvement for precise affordance reasoning in 3DGS scenes, suggesting its superiority in capturing fine-grained affordance detail.
However, this method targets at a specific and controlled task prototype, where each scene consists of a single instances and each instruction requires just a single atomic action for execution.

Due to the inherent high-level succinctness of real-world instructions, 3D affordance reasoning requires both (i) the composition of ordered primitive instructions (each corresponding to a primitive action), and (ii) the dynamic shift of actionable regions across object instances in complex scenes.
As illustrated in Fig. \ref{fig:teaser}, an instruction such as \textit{operating a microwave to warm up food in a bowl} necessitates multiple interdependent actions across three distinct actionable regions from different instances---a composite capability beyond current methods due to their constrained task prototypes.
This reveals a critical gap in 3D affordance reasoning: the absence of a task formulation for sequential interaction in cluttered scenes. 
To bridge this gap, we introduce the \textbf{Scene-level Sequential 3D Gaussian Affordance Reasoning} task---a novel prototype designed for long-horizon and succinct instructions in complex environments with multiple interactive regions and distractor objects, fundamentally advancing beyond prior object-centric, single-action approaches.

To facilitate research in this new direction, we introduce \textbf{SeqAffordSplat}, the first comprehensive benchmark designed for long-horizon, scene-wide affordance reasoning on 3DGS. 
The benchmark features a new large-scale dataset containing over 1,800+ complex scenes, 14,000+ affordance masks, and 8,000+ sequential instructions. 
To establish a complete evaluation framework, we complement the dataset with a suite of novel sequential metrics—\textbf{sIoU}, \textbf{sAUC}, \textbf{sSIM}, and \textbf{sMAE}—tailored to holistically assess performance on multi-step tasks.

To handle the shift in task prototype, we introduce \textbf{SeqSplatNet}, the first framework designed to solve this new scene-level sequential task. Our approach uniquely integrates the hierarchical planning capabilities of Large Language Models (LLMs) with the rich representational power of 3DGS in a unified, end-to-end architecture. Given a complex, long-horizon instruction, our model reasons about the user's intent and grounds a sequence of actionable affordance masks directly onto the 3DGS scene representation. 
To handle complex scene geometry, we design a self-supervised pre-training strategy, Conditional Geometric Reconstruction, where the model learns to reconstruct complete affordance regions from partial geometric observations, thereby building a robust geometric prior. Furthermore, to resolve semantic ambiguities, we design a Semantic Feature Injection mechanism that lifts rich semantic features from a frozen 2D Vision Foundation Model (VFM)~\cite{oquab2023dinov2,radford2021learning} via multi-view rendering and fuses them into the 3D decoder at multiple scales. This unified approach bridges the critical gap between high-level task planning and low-level, fine-grained 3D perception in complex environments.
Our main contributions are summarized as follows:

\begin{itemize}
    \item  We introduce the new task of Sequential 3D Gaussian Affordance Reasoning and develop SeqAffordSplat, the first large-scale benchmark with over 1,800+ 3DGS scenes and 14,000+ ground-truth affordance segmentations.
    \item We propose SeqSplatNet, the first framework to unify high-fidelity 3DGS representation, long-horizon sequential planning, and complex scene-level understanding.
   
    \item We demonstrate through extensive experiments that our approach achieves a 14.1\% performance improvement over sequential baselines on our challenging new benchmark.
  
\end{itemize}

\section{Related Work}

\subsection{Affordance Learning}

Affordance Learning focuses on identifying interactive regions in a scene, driven by either a closed-set of action types or open-vocabulary language instructions.
This originates from 2D image affordance segmentation, where each pixel is assigned to a predefined affordance category~\cite{do2018affordancenet,roy2016multi}.
To generalize to unseen affordance types, recent approaches integrate Vision-Language Models (VLMs), aligning language instructions with visual affordances~\cite{chen2025maskprompt,li2024one,qian2024affordancellm}.
However, these image-based approaches intrinsically lack the ability to capture explicit 3D spatial information---a critical requirement for robotic manipulation applications.

To overcome the limitations of image-based perception, researchers turns to 3D representations for acurrate gemoetry awareness in 3D space. 
Pioneering studies ~\cite{deng20213d,xu2022partafford,mo2022o2o} established benchmarks for affordance segmentation on 3D point clouds for predefined affordance type sets. 
This foundation later has evolved toward open-vocabulary affordance reasoning, where models identify actionable regions in response to language instructions by leveraging the cross-modality reasoning capability of foundational models~\cite{li2024laso,xu2022partafford,shao2025great,lu2025geal}. 
Despite this evolution, these approaches still struggle with long-horizon reasoning, which stems from their constrained task prototype that each instruction envolves a single action.

To address this limitation, we highlight sequential affordance reasoning as a more practical task prototype, where each instruction maps to a sequence of atomic affordances.
Compared with the contemporary SeqAfford~\cite{yu2025seqafford}, which is capable of generating sequential affordance masks on 3D cloud points, our SeqSplatNet overcomes its inefficiency in localizing fine-grained actionable regions through high-fidelity 3DGS representation.

\subsection{Affordance Learning on 3DGS}

Beyond 3D point clouds, 3DGS offers a more expressive representation through its explicit point-based structure and real-time rendering capabilities~\cite{kerbl20233d}.
These characteristics facilitate the development of various embodied AI systems~\cite{zheng2024gaussiangrasper,shorinwa2024splat,lu2024manigaussian}, as 3DGS enables both instantaneous environmental perception and direct association of semantic information with spatially precise geometric locations.

In affordance learning, 3DAffordSplat~\cite{wei20253daffordsplat} established the first large-scale benchmark for affordance reasoning using 3DGS. While notable for its substantial instance volume and diverse affordance coverage, this benchmark adopts a constrained task prototype where each instruction maps exclusively to a single discrete affordance mask, inherently omitting sequential reasoning pathways essential for complex manipulation scenarios.

In summary, existing approaches remain confined to either sequential reasoning on sparse point clouds with inadequate localization fidelity or single-step interactions on 3DGS without action sequencing capability. 
To the best of our knowledge, we pioneer the first framework for scene-level affordance learning in 3DGS that simultaneously achieves fine-grained affordance localization and sequential reasoning.

\section{Task Definition and Dataset}

\begin{figure*}
    \centering
    \includegraphics[width=0.85\linewidth]{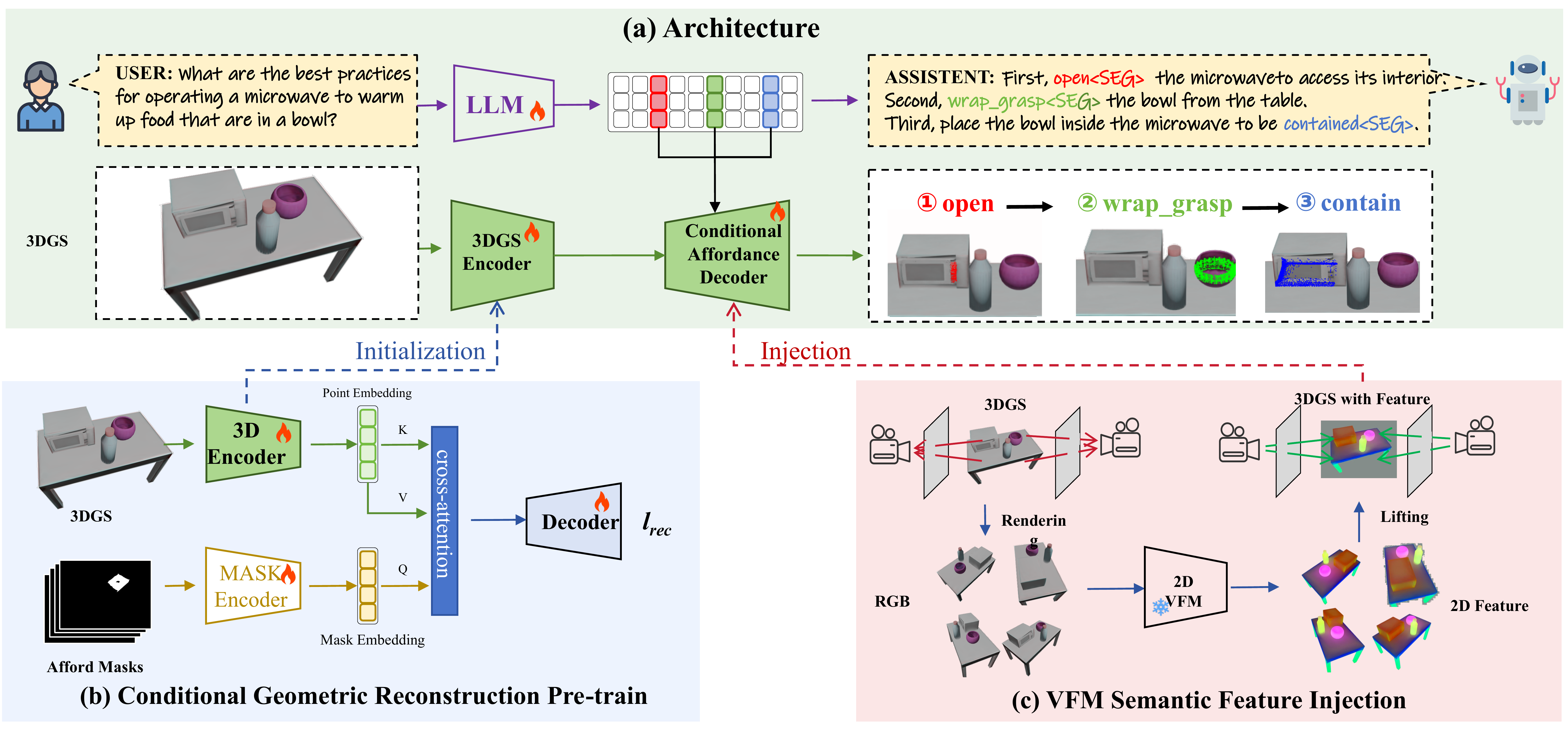}
    \caption{An overview of the proposed \textbf{SeqSplatNet} architecture. The architecture comprises four main components: a Large Language Model, a 3DGS Encoder with Conditional Geometric Reconstruction Pre-train, and a Conditional Affordance Decoder with VFM Semantic Feature Injection. }
    \label{fig:pipline}
\end{figure*}

\subsection{Task Definition}

Scene-Level Sequential 3D Gaussian Affordance Reasoning presents a challenging task requiring the identification of step-wise affordance regions in a complex 3DGS scene containing multiple interactive instance parts, in accordance with a succinct input instruction that outlines a composite process of multiple ordered primitive actions.

Specifically, consider a 3D scene represented by a 3DGS model $\mathcal{G}$ comprising $N$ Gaussian primitives, $\mathcal{G} = \{G_i\}_{i=1}^N$. 
Each primitive $G_i$ is parameterized by its position, opacity, scale, rotation, and spherical harmonics coefficients.
Given a succinct instruction $Q_{inst}$, this task aims to predict an ordered sequence of $T$ binary affordance masks $\mathcal{M}=(M_{1}, M_{2},...,M_{T})$.
Each mask $M_t \in \{0, 1\}^N$ identifies the subset of Gaussians that constitute the functional region for the $t$-th atomic instruction, which is implicitly defined by the instruction's action plan, as illustrated in Fig. \ref{fig:teaser}. 
The objective of this task is to find an optimal mapping $F$ that satisfies
\begin{equation}
     \mathcal{M} = F(Q_{inst}, \mathcal{G}).
\end{equation}
This formulation extends the traditional affordance segmentation paradigm from identifying \textit{what} affordances exist to reasoning about \textit{in what order} they must be actualized to fulfill a user's complex intent.

\subsection{SeqAffordSplat Dataset Collection}
To facilitate research into long-horizon, scene-level affordance reasoning, we introduce SeqAffordSplat, the first large-scale benchmark designed to evaluate sequential affordance grounding directly on 3DGS representations.  The construction of the SeqAffordSplat benchmark is a two-stage process meticulously designed to ensure high fidelity and ecological validity across its two core components: the 3D scene geometry, and the sequential, language-grounded instructions.

\subsubsection{Step 1: 3DGS Data Collection.}
The foundation of our benchmark lies in the quality and complexity of its 3D environments. To properly evaluate long-horizon reasoning, which often involves interactions among multiple objects, single-object models are insufficient.To generate realistic environments, we manually composed scenes by positioning multiple objects from the 3D-AffordanceNet~\cite{wei20253daffordsplat} dataset using geometric transformations, including translation, rotation, and scaling, to emulate plausible real-world scenarios.

\subsubsection{Step 2: Instruction and Affordance Annotation.}
Rather than annotating from scratch, we transfer affordance labels from established benchmarks, primarily 3D-AffordanceNet ~\cite{wei20253daffordsplat}, which provides dense, point-wise affordance labels for thousands of object shapes. The transfer is performed via a semi-automated pipeline: we programmatically match object instances in our scenes with annotated categories in 3DAffordSplat~\cite{wei20253daffordsplat}, project the point-wise labels onto our 3DGS representation, and then use a custom 3D annotation tool for manual verification. For a sequential instruction, the ground truth is stored as an ordered list of affordance masks, explicitly encoding the temporal and causal order required for the task.
To generate a large and diverse set of long-horizon instructions, we utilize the multimodal large language model (MLLM) GPT-4o~\cite{achiam2023gpt} inspired by ~\cite{yu2025seqafford}. We employ a sophisticated prompt engineering strategy that provides the LLM with rich context for each scene, including Visual Context, Textual Context, Role Prompting and Goal Specification. The generated instruction-sequence pairs undergo a final human-in-the-loop curation process to ensure they are logical, physically possible, and correspond to available affordances. Additional details are provided in the supplementary materials.

As shown in \ref{tab:dataset}, the final benchmark contains 1800+ unique 3DGS scenes, annotated with over 14,000+ distinct affordance masks across 21 object categories and 18 affordance types. The language component features approximately 8000+ instructions. A key characteristic is its focus on long-horizon tasks.

\subsection{Evaluation Configurations and Metrics}
\noindent \textbf{Experimental Settings.} Inspired by the evaluation settings in prior work~\cite{yu2025seqafford,li2024laso}, we establish three distinct configurations to comprehensively evaluate our method:
\begin{itemize}
    \item \textit{Single}: Evaluates the model's ability to predict individual, unordered affordance regions.
    \item \textit{Sequential (with gt seq)}: Assesses affordance grounding accuracy given a ground-truth action sequence.
    \item \textit{Sequential}: Tests the full task, where the model must infer and execute the entire action sequence from a single high-level instruction.
\end{itemize}

\noindent \textbf{Evaluation Metrics.} For single-step prediction, we adopt standard metrics \textbf{mIoU}, \textbf{AUC}, \textbf{SIM}, \textbf{MAE} to ensure a fair comparison with prior works like LASO~\cite{li2024laso}. 
For sequential task , we introduce a suite of sequential metrics: \textbf{sIoU}, \textbf{sAUC}, \textbf{sSIM}, and \textbf{sMAE}. The calculation is straightforward: we first align the predicted and ground-truth sequences to the same length by padding the shorter sequence with empty frames. 
This approach is reasonable because it inherently penalizes discrepancies in sequence length.  Additional details are provided in the supplementary materials.

\section{SeqSplatNet}

\subsection{Architecture}

Our SeqSplatNet features an end-to-end architecture that directly maps language instructions to sequential 3D affordance masks. 
Through an autoregressive process, the model generates interleaved language tokens and special \texttt{<SEG>} tokens, where each \texttt{<SEG>} emission dynamically triggers the affordance decoder to produce a 3D affordance mask. 
This design inherently unifies task planning and localization by embedding action sequencing within the generative process, eliminating explicit hierarchical decomposition.

As illustrated in Fig.~\ref{fig:pipline}, our SeqSplatNet comprises three core components: a 3DGS Encoder, a Large Language Model (LLM) and a Conditional Affordance Decoder, collectively constituting our base model. 
We augment this framework with two key enhancements: Conditional Geometric Reconstruction Pre-train for improved 3DGS Encoder initialization, and VFM Semantic Feature Injection to enrich geometric representations with semantic knowledge extracted from 2D VFMs.

\textbf{3DGS Encoder.} 
We adopt a PointNet-based encoder ~\cite{qi2017pointnet++} to extract geometric information from a 3DGS scene $\mathcal{G}$. 
Consistent with 3DAffordSplat~\cite{wei20253daffordsplat}, our encoder processes the geometric attributes (\texttt{position}, \texttt{rotation} and \texttt{scale}) of Gaussian primitives in $\mathcal{G}$, generating point-wise geometric features $F_{\text{geo}} \in \mathbb{R}^{N \times d}$.

\textbf{LLM.} 
Our LLM serves as the central reasoning engine, processing an input instruction $Q_{\text{instr}}$ to autoregressively generate a primitive instruction sequence. 
Inspired by recent advancements in Multimodal LLM~\cite{li2024laso, wei20253daffordsplat}, we augment the token vocabulary with a special token \texttt{<SEG>}.
Within the interleaved sequence of language tokens and \texttt{<SEG>} tokens, each \texttt{<SEG>} simultaneously activates affordance mask decoding for its associated primitive instruction and provides a dynamic instruction vector $h_{\text{seg}} \in \mathbb{R}^d$ derived from its hidden state. 
Benifiting from the masked attention mechanism in LLM, this vector effectively encodes the contextual dependencies from $Q_{\text{instr}}$ and its preceding primitives.

\textbf{Conditional Affordance Decoder.} 
This decoder generates the affordance mask conditioned on each obtained dynamic instruction vector $h_{\text{seg}}$. 
Built upon recent query-based segmentation paradigm~\cite{cheng2021per}, it employs each LLM-derived dynamic instruction vector $h_{\text{seg}}$ as a latent query to decode its corresponding 3D affordance mask $M_t$ from the encoded geometric feature and injected semantic information (as detailed subsequently).

This tight integration of reasoning and perception within a unified autoregressive framework enables end-to-end sequential reasoning in complex 3DGS scenes with our SeqSplatNet.

\subsubsection{End-to-end Training of SeqSplatNet.}

Our SeqSplatNet aims to generate fine-grained affordance masks associated for accurately reasoned primitive instruction sequences for succinct input instructions.
To this end, its overall loss summarizes both the language-modeling misalignments $\mathcal{L}_{\text{lang}}$ and the affordance segmentation errors $\mathcal{L}_{\text{mask}}$, 
\begin{equation}
    \mathcal{L}_{\text{total}} = \mathcal{L}_{\text{lang}} + \lambda_{\text{mask}} \sum_{t=1}^T \mathcal{L}_{\text{mask}},
\end{equation}
where $\lambda_{\text{mask}}$ is a balancing hyperparameter. 
In this work, we adopt the standard autoregressive cross-entropy loss over the predicted token sequence for $\mathcal{L}_{\text{lang}}$.
The segmentation loss $\mathcal{L}_{\text{mask}}$, activated at each \texttt{<SEG>} token, is a combination of Binary Cross-Entropy (BCE) and Dice losses to ensure both pixel-wise accuracy and structural similarity:
\begin{equation}
\mathcal{L}_{\text{mask}} = \mathcal{L}_{\text{BCE}}(\hat{M}_t, M_t^{\text{gt}}) + \mathcal{L}_{\text{Dice}}(\hat{M}_t, M_t^{\text{gt}}),
\end{equation}
where $\hat{M}_t$ and $M_t^{\text{gt}}$ are the predicted and ground-truth masks for step $t$, respectively.

\begin{table*}[htbp]
  \centering
  \small
  \caption{Results on SeqAffordSplat dataset}
    \begin{tabular}{l|l|c|cccc}
    \toprule
    \multicolumn{1}{l|}{Main results} & Method & Source & \textit{mIoU/sIoU↑} & \textit{AUC/sAUC↑} & \textit{SIM/sSIM↑} & \textit{MAE/sMAE↓} \\
    \midrule
    \multirow{4}{*}{\textbf{Single}} & 3DAffordSplat & ACM MM'25 & 30.5  & 92.7  & 0.395  & 0.065  \\
          & PointRefer & CVPR'24 & 31.3  & 92.1  & 0.411  & 0.055  \\
          & IAGNet & ICCV'23 & 17.6  & 85.2  & 0.328  & 0.056  \\
          & \textbf{OURS} & - & \textbf{37.0} & \textbf{94.0} & \textbf{0.470} & \textbf{0.049} \\
    \midrule
    \multirow{4}{*}{\makecell[l]{\textbf{Sequential}\\\textbf{(with GT seq)}}} & 3DAffordSplat & ACM MM'25 & 26.1 & 91.2  & 0.343  & 0.072  \\
          & PointRefer & CVPR'24 & 30.3  & 91.2  & 0.418  & 0.055  \\
          & IAGNet & ICCV'23 & 13.9  & 88.0  & 0.325  & 0.062  \\
          & \textbf{OURS} & - & \textbf{36.0} & \textbf{95.6} & \textbf{0.457} & \textbf{0.036} \\
    \midrule
    \multirow{2}{*}{\textbf{Sequential}} & SeqAfford & CVPR'25 &  12.1 &  73.0 & 0.122  &  0.230 \\
          & \textbf{OURS} & - & \textbf{26.2} & \textbf{80.6} & \textbf{0.312} & \textbf{0.132} \\
    \bottomrule
    \end{tabular}%
  \label{tab:SeqAffordSplat}%
\end{table*}%

\subsection{Conditional Geometric Reconstruction Pre-train}
Obaining an effective 3DGS encoder from scratch for complex scene-level 3DGS is challenging, due to its reliance on a huge amount of annotated samples. 
To tackle this issue, we propose a pre-training strategy for providing improved initialization to our perception modules.
Our core idea is to instill a  geometric prior into the 3DGS encoder by tasking it with reconstructing a spatial affordance region conditioned solely on an abstract semantic embedding.

Specifically, given a 3DGS scene $\mathcal{G}$ and an affordance mask $M^{\text{gt}} \in \{0, 1\}^N$, our architecture employs a dual-encoder design. 
The 3DGS encoder $\Phi_{\text{enc}}$ projects the scene $\mathcal{G}$ into per-point feature embeddings $F_{\text{geo}}$, and the mask encoder $\Phi_{\text{mask}}$ maps the mask $M^{\text{gt}}$ into a single conditional embedding $e_{\text{mask}}$ that represents the abstract affordance concept. 
The reconstruction is then conditioned on this embedding. Specifically, $e_{\text{mask}}$ serves as a query to attend to the per-point geometric features $F_{\text{geo}}$, producing a fused representation $F_{\text{fused}}$. A decoder, $\Phi_{\text{dec}}$, then processes these features to reconstruct the final mask $\hat{M}$:
\begin{align}
    F_{\text{fused}} &= \text{Attention}(Q=e_{\text{mask}}, K=F_{\text{geo}}, V=F_{\text{geo}}) \\
    \hat{M} &= \Phi_{\text{dec}}(F_{\text{fused}})
\end{align}
This pre-training task compels the network to learn a powerful mapping from an abstract semantic concept to its corresponding spatial geometry. By explicitly conditioning the reconstruction on the mask embedding, the model learns a disentangled representation that separates the affordance concept from the geometric structure, providing a superior initialization for the downstream task.

\subsection{VFM Semantic Feature Injection}
Interpreting nuanced language instructions to identify affordances requires a deep semantic understanding that pure geometric representations cannot provide for complex scenes. 
To bridge this gap, we leverage the high-fidelity rendering capability of 3DGS to inject potent semantic knowledge from pre-trained 2D Vision Foundation Models (VFM). 

For a scene represented by a set of $n$ 3D Gaussians, we first generate $m$ multi-view 2D feature maps $\{F^{(v)}\}_{v=1}^m$. Each feature map is obtained by processing a rendered RGB image $I^{(v)}$ with a frozen, pre-trained VFM, $\Psi_{\text{VFM}}$ (e.g., DINOv2~\cite{oquab2023dinov2},CLIP~\cite{radford2021learning}):
\begin{equation}
    F^{(v)} = \Psi_{\text{VFM}}(I^{(v)}) \in \mathbb{R}^{H \times W \times d_{\text{sem}}}
\end{equation}
To lift these 2D features into the 3D space, we employ a learning-free aggregation process akin to an inverse rendering operation following the learning-free lifting paradigm~\cite{marrie2024ludvig} . This approach correctly handles the contribution of multiple Gaussians to each rendered pixel via alpha-blending. The semantic feature vector $f_i^{\text{sem}}$ for each Gaussian $i$ is computed as a weighted average of all the 2D pixel features it influences across all views. The weight for each pixel feature $F_{p}^{(v)}$ (from view $v$ at pixel $p$) is its rendering weight $w_i(v, p)$, which represents the influence of Gaussian $i$ on that pixel. The lifted feature is defined as:
\begin{equation}
    f_i^{\text{sem}} = \frac{\sum_{(v, p) \in \mathcal{S}_i} w_i(v, p) F_{p}^{(v)}}{\sum_{(v, p) \in \mathcal{S}_i} w_i(v, p)}
    \label{eq:uplifting}
\end{equation}
where $\mathcal{S}_i$ is the set of all view-pixel pairs $(v, p)$ that Gaussian $i$ contributes to. This aggregation method inverts the rendering process to produce a semantically-rich feature bank $F_{\text{sem}} \in \mathbb{R}^{n \times d_{\text{sem}}}$ attached to the Gaussians.

The the lifted semantic features $F_{\text{sem}}$ are injected into the Conditional Affordance Decoder at multiple scales using additive fusion, this multi-scale strategy enhances semantic consistency in segmentation by informing the decoding process at all levels of granularity, from coarse to fine-grained.

\section{Experiments}

\subsection{Experimental Settings}

\noindent\textbf{Baseline Models.} Since our method is the first sequence reasoning approach based on 3DGS, for a fair comparison, we selected the following baselines: 3DAffordSplat~\cite{wei20253daffordsplat}, a single-step reasoning method based on 3DGS; PointRefer~\cite{li2024laso} and IAGNet~\cite{yang2023grounding}, single-step methods based on point clouds; and SeqAfford~\cite{yu2025seqafford}, a sequence reasoning method based on point clouds. We evaluate these methods under various settings.

\noindent\textbf{Implementation Details.} For our primary results, we chose the Qwen-3-0.6B~\cite{yang2025qwen3} as the LLM, formatting inputs with its official chat template. The model was fine-tuned for our task using Low-Rank Adaptation (LoRA)~\cite{hu2022lora}. 
Both geometric and semantic feature dimensions projected set to 512. The model was first pre-trained for 10 epochs at a learning rate of $1 \times 10^{-4}$. It was then trained for 50 epochs, where all non-LLM parameters used a decaying learning rate of $1 \times 10^{-4}$. For Qwen's LoRA fine-tuning, we targeted the \texttt{q\_proj}, \texttt{k\_proj}, \texttt{v\_proj}, and \texttt{lm\_head} layers with a rank of 8 and an alpha of 16. 
The Adam optimizer~\cite{kingma2014adam} with weight decay  was used for both training stages. Experiments were conducted on 8 GeForce RTX 3090 GPUs.


\subsection{Results on SeqAffordSplat Dataset}
We performed the Sequential 3D Gaussian Affordance Reasoning task on the SeqAffordSplat Dataset. As outlined in the Dataset section, this task is categorized into three distinct settings based on the nature of the instructions: \textit{Single}, \textit{Sequential (with ground-truth sequence)}, and end-to-end \textit{Sequential}. The main results are presented in Table~\ref{tab:SeqAffordSplat}.

\noindent \textbf{Single.} In the single-step setting, models predict an affordance mask for a single, explicit instruction. Our method achieves a state-of-the-art performance with an \textit{mIoU} of 37.0 and an \textit{AUC} of 94.0. This represents a significant improvement of 5.7 points in \textit{mIoU} over the strongest point-cloud-based baseline, PointRefer (31.3), and a 6.5-point improvement over the 3DGS-based baseline, 3DAffordSplat (30.5). Notably, the performance of 3DAffordSplat is slightly lower than that of PointRefer. A possible reason is that 3DAffordSplat treats the 3D Gaussians simply as a point cloud enriched with features like opacity, scale, and rotation, without fully leveraging the high-fidelity rendering capabilities inherent to 3DGS. This demonstrates the superior capability of our architecture in understanding and grounding instructions even in non-sequential scenarios.

\begin{figure*}
    \centering
    \includegraphics[width=1.0\linewidth]{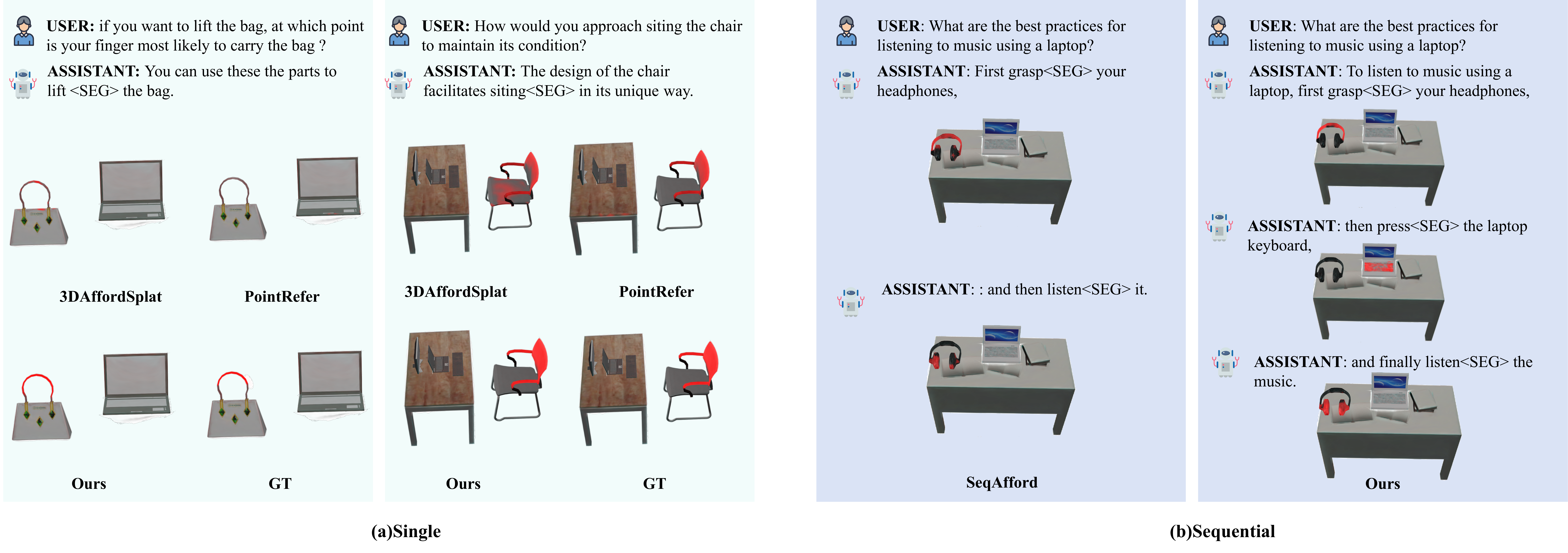}
    \caption{Visual Results of our proposed methods. }
    \label{fig:visual}
\end{figure*}

\noindent \textbf{Sequential(with GT seq).} This setting evaluates the model's ability to ground affordances given a ground-truth sequence of sub-instructions. This isolates the performance of the perception module from the language reasoning component. Our method continues to outperform all baselines, achieving an \textit{sIoU} of 36.0. This is 5.7 points higher than the next-best baseline, PointRefer, indicating that our conditional decoder excels at accurately interpreting specific sub-tasks and generating precise affordance masks.

\noindent \textbf{Sequential.} This is the full end-to-end task, where the model must reason about a complex instruction and generate the entire sequence of affordance masks. Since other baselines do not support end-to-end sequential reasoning, we compare our method against SeqAfford~\cite{yu2025seqafford}, the only available baseline for this task, which operates on point clouds. In this challenging setting, our method demonstrates a remarkable improvement. We achieve an \textit{sIoU} of 26.2, which is more than double the performance of SeqAfford (12.1). This substantial gain of 14.1 points in \textit{sIoU} underscores the effectiveness of our integrated reasoning and perception framework, which leverages the LLM to decompose tasks and the decoder to ground them in the 3DGS representation.

\begin{table}[t]
  \centering
  \small
  \caption{Results on 3DAffordSplat dataset}
    \begin{tabular}{l|cccc}
    \toprule
    Method & \textit{mIoU↑} & \textit{AUC↑} & \textit{SIM↑} & \textit{MAE↓} \\
    \midrule
    3DAffordSplat & 30.3  & 83.9  & 0.440  & 0.210  \\
    IAGNet & 14.6  & 56.7  & 0.350  & 0.410  \\
    PointRefer & 18.4  & 78.5  & 0.430  & 0.200  \\
    \textbf{OURS} & \textbf{40.2} & \textbf{89.3} & \textbf{0.530} & \textbf{0.169} \\
    \bottomrule
    \end{tabular}%
  \label{tab:3DAffordSplat}%
\end{table}%

\subsection{Results on 3DaffordSplat Datasets} To validate the generalization capability of our approach, we also evaluated it on the existing 3DAffordSplat dataset~\cite{wei20253daffordsplat}, which focuses on single-step affordance reasoning on 3D Gaussian data. As shown in Table~\ref{tab:3DAffordSplat}, our method achieves an \textit{mIoU} of 40.2, significantly outperforming all prior methods. This result is 9.9 points higher than the original 3DAffordSplat benchmark, confirming that the architectural designs and training strategies proposed in our work are robust and effective beyond our new sequential task, setting a new state-of-the-art on this established benchmark as well.

\begin{table}[htbp]
  \centering
  \small
  \caption{Ablation Study of Main Components}
    \begin{tabular}{ll|rrrr}
    \toprule
    \multicolumn{2}{c|}{Component} & \multirow{2}{*}{\textit{sIoU↑}} & \multirow{2}{*}{\textit{sAUC↑}} & \multirow{2}{*}{\textit{sSIM↑}} & \multirow{2}{*}{\textit{sMAE↓}} \\
    \cmidrule(lr){1-2} 
    Pretrain & Feature &       &       &       &  \\
    \midrule
    $\times$      & $\times$      & 20.3  & 76.3  & 0.229 & 0.169 \\
    \checkmark &  $\times$     & 24.1  & 78.5  & 0.302 & 0.141 \\
    \checkmark    & CLIP  & 24.2  & 79.1  & 0.290 & 0.141 \\
    \checkmark & DINO v2   & \textbf{26.2} & \textbf{80.6} & \textbf{0.312} & \textbf{0.132} \\
    \bottomrule
    \end{tabular}%
  \label{tab:Components}%
\end{table}%

\begin{table}[htbp]
  \centering
  \small
  \caption{Ablation Study of LLM backbones}
    \begin{tabular}{l|cccc}
    \toprule
    \multicolumn{1}{l|}{LLM} & \multicolumn{1}{c}{\textit{sIoU↑}} & \multicolumn{1}{c}{\textit{sAUC↑}} & \multicolumn{1}{c}{\textit{sSIM↑}} & \multicolumn{1}{c}{\textit{sMAE↓}} \\
    \midrule
    GPT2-small(0.1B)  &12.1        &43.9        &0.156        &0.488  \\
    \textbf{Qwen3-0.6B}   &\textbf{26.2} &\textbf{80.6} & \textbf{0.312} &\textbf{0.132} \\
    Qwen3-1.7B  &26.4     &79.5      &0.291       &0.147  \\
    Qwen3-8B  & 24.2    &78.6      &0.285        & 0.148 \\
    \bottomrule
    \end{tabular}%
  \label{tab:backbones}%
\end{table}%

\subsection{Ablation Study}
We conducted ablation studies to analyze the contribution of each key component in our framework on \textit{Sequential} task. 

\noindent \textbf{Effects of Different Components.} The effectiveness of our main components is validated in Table~\ref{tab:Components}. Our Conditional Geometric Reconstruction pre-training provides a substantial performance boost over a baseline model, improving the \textit{sIoU} from 20.3 to 24.1. The subsequent injection of rich semantic features from DINOv2 further lifts the performance to 26.2 \textit{sIoU}. This demonstrates that both our pre-training strategy and semantic feature fusion are critical to the model's success.

\noindent \textbf{Effects of Different LLM Encoders.} We investigated the impact of the LLM backbone on reasoning performance, with results shown in Table~\ref{tab:backbones}. The results demonstrate that the choice of LLM is critical for the sequential reasoning task. The GPT2-small (0.1B)\cite{radford2019language} model serves as a baseline, achieving a mere 12.1 \textit{sIoU}, which underscores the necessity of a more capable language model. Among the Qwen3 series, our primary model, Qwen3-0.6B, performs very competitively with an \textit{sIoU} of 26.2, while also achieving the best \textit{sAUC} (80.6) and \textit{sSIM} (0.312) scores. Interestingly, the largest model tested, Qwen3-8B, shows a performance degradation with an \textit{sIoU} of 24.2. This suggests that simply increasing parameter count does not guarantee better performance on this task. Consequently, we selected Qwen3-0.6B for our main experiments as it offers an excellent trade-off between high performance across multiple metrics and model efficiency.

\subsection{Qualitative Results}
Figure~\ref{fig:visual} presents a qualitative comparison of our method against baselines. In single-step scenarios (a), our model demonstrates superior precision. For example, it correctly identifies the specific liftable part of an bag  based on a nuanced instruction, while competing methods segment the incorrect region. More critically, for the sequential task (b), our model successfully decomposes a high-level command (``listen to music using a laptop") into a logical, multi-step sequence of affordances. In contrast, the baseline method fails to generate a coherent plan, visually validating our framework's advanced reasoning and grounding capabilities for complex, long-horizon tasks.

\section{Conclusion}
\input{6_conclusion}
\bibliography{aaai2026}

\end{document}

%% file: 6_conclusion.tex
In this paper, we advance 3D affordance reasoning from single-step, object-centric interactions to complex, sequential tasks at the scene level. We introduced SeqSplatNet, the first framework to unify a causal language model with a high-fidelity 3DGS representation for this new paradigm. Bolstered by novel geometric pre-training and semantic feature injection techniques, our method establishes a new state-of-the-art, outperforming the prior sequential baseline by 14.1\% on our newly proposed SeqAffordSplat benchmark. This work provides a critical foundation for developing more capable embodied agents that can understand and execute long-horizon instructions in complex environments. 